\documentclass[11pt,letterpaper]{article}
\usepackage{emnlp2017}
\usepackage{times}
\usepackage{latexsym}
\usepackage{booktabs} 
\usepackage{multirow}
\usepackage{arydshln}
\usepackage{graphicx}

\newcommand{\enotesoff}{\long\gdef\enote##1##2{}}

\enotesoff

\emnlpfinalcopy

\def\emnlppaperid{***}

\newcommand{\stag}{\textsc{Stag}}
\newcommand{\upos}{\textsc{Pos}}
\newcommand{\morph}{\textsc{Morph}}

\title{A General-Purpose Tagger with Convolutional Neural Networks}

\author{Xiang Yu \and Agnieszka Fale\'{n}ska \and Ngoc Thang Vu \\
  Institut f\"ur Maschinelle Sprachverarbeitung\\
  Universit\"at Stuttgart\\
  {\tt \{xiangyu,falensaa,thangvu\}@ims.uni-stuttgart.de} }

\date{}

\begin{document}

\maketitle

\begin{abstract}
  We present a general-purpose tagger based on convolutional neural networks (CNN), used for both composing word vectors and encoding context information. The CNN tagger is robust across different tagging tasks: without task-specific tuning of hyper-parameters, it achieves state-of-the-art results in part-of-speech tagging, morphological tagging and supertagging. The CNN tagger is also robust against the out-of-vocabulary problem, it performs well on artificially unnormalized texts.
\end{abstract}

\section{Introduction}

Recently, character composition models have shown great success in many NLP tasks, mainly because of their robustness in dealing with out-of-vocabulary (OOV) words by capturing sub-word informations. Among the character composition models, bidirectional long short-term memory (LSTM) models and convolutional neural networks (CNN) are widely applied in many tasks, e.g. part-of-speech (POS) tagging \citep{Santos:2014, Plank:2016}, named entity recognition \citep{Santos:2015}, language modeling \citep{Ling:2015, Kim:2016}, machine translation \citep{Costa:2016} and dependency parsing \citep{Ballesteros:2015, Yu:2017}. 


In this paper, we present a state-of-the-art general-purpose tagger that uses CNNs both to compose word representations from characters and to encode context information for tagging.\footnote{We will release the code in the camera-ready version.} We show that the CNN model is more capable than the LSTM model for both functions, and more stable for unseen or unnormalized words, which is the main benefit of character composition models.

\newcite{Yu:2017} compared the performance of CNN and LSTM as character composition model for dependency parsing, and concluded that CNN performs better than LSTM. In this paper, we show that this is also the case for POS tagging. Furthermore, we extend the scope to morphological tagging and supertagging, in which the tag set is much larger and long-distance dependencies between words are more important.



In these three tagging tasks, we compare our tagger with the \textbf{bilstm-aux} tagger \citep{Plank:2016} and the CRF-based morphological tagger \textbf{MarMot} \citep{Mueller:2013}. The CNN tagger shows robust performance accross the three tasks, and achieves the highest average accuracy in all tasks. It (significantly) outperforms LSTM in morphological tagging, and outperforms both baselines in supertagging by a large margin.


To test the robustness of the taggers against the OOV problem, we also conduct experiments using artificially constructed unnormalized text by corrupting words in the normal dev set. Again, the CNN tagger outperforms the two baselines by a very large margin.



Therefore we conclude that our CNN tagger is a robust state-of-the-art general-purpose tagger that can effectively compose word representation from characters and encode context information. 


\section{Model}
\begin{figure}[t]
\includegraphics[width=.48\textwidth]{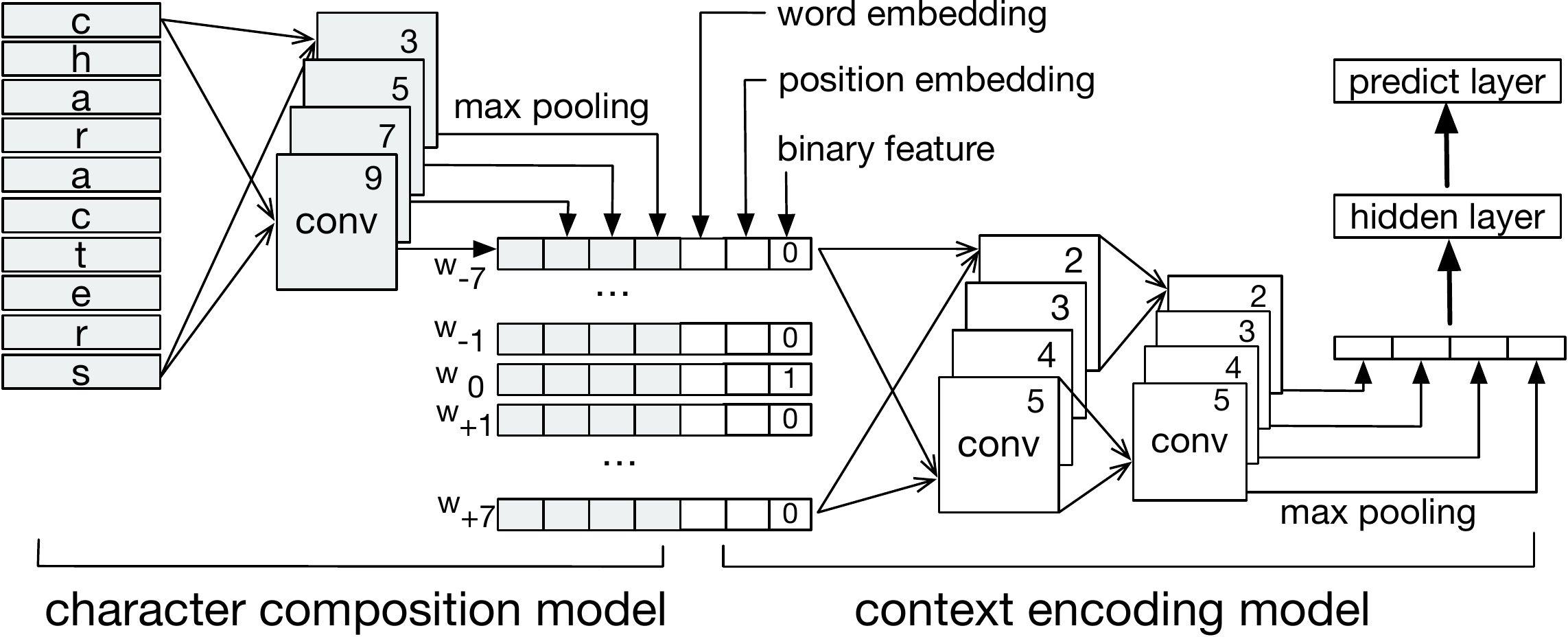}
\caption{Diagram of the CNN tagger.\label{fig:cnn}}
\end{figure}

Our proposed CNN tagger has two main components: the character composition model and the context encoding model. Both components are essentially CNN models, capturing different levels of information: the first CNN captures morphological information from character n-grams, the second one captures contextual information from word n-grams. Figure~\ref{fig:cnn} shows a diagram of both models of the tagger.

\subsection{Character Composition Model}
The character composition model is similar to \newcite{Yu:2017}, where several convolution filters are used to capture character n-grams of different sizes. The outputs of each convolution filter are fed through a max pooling layer, and the pooling outputs are concatenated to represent the word. 

\subsection{Context Encoding Model}
The context encoding model captures the context information of the target word by scanning through the word representations of its context window. The word representation could be only word embeddings ($\vec{w}$), only composed vectors ($\vec{c}$) or the concatenation of both ($\vec{w}+\vec{c}$)

 A context window consists of N words to both sides of the target word and the target word itself. 
To indicate the target word, we concatenate a binary feature to each of the word representations with 1 indicating the target and 0 otherwise, similar to \newcite{Vu:2016}. 
Additional to the binary feature, we also concatenate a position embedding to encode the relative position of each context word, similar to \newcite{Gehring:2017}.

\subsection{Hyper-parameters}\label{hyper}
For the character composition model, we take a fixed input size of 32 characters for each word, with padding on both sides or cutting from the middle if needed. We apply four convolution filters with sizes of 3, 5, 7, and 9. Each filter has an output channel of 25 dimensions, thus the composed vector is 100-dimensional. We apply Gaussian noise with standard deviation of 0.1 is applied on the composed vector during training. 

For the context encoding model, we take a context window of 15 (7 words to both sides of the target word) as input and predict the tag of the target word. 
We also apply four convolution filters with sizes of 2, 3, 4 and 5, each filter is stacked by another filter with the same size, and the output has 128 dimensions, thus the context representation is 512-dimensional. We apply one 512-dimensional hidden layer with ReLU non-linearity before the prediction layer. We apply dropout with probability of 0.1 after the hidden layer during training.

The model is trained with averaged stochastic gradient descent with a learning rate of 0.1, momentum of 0.9 and mini-batch size of 100. We apply L2 regularization with a rate of $10^{-5}$ on all the parameters of the network except the embeddings.

\section{Experiments}
\subsection{Data}
We use treebanks from version 1.2 of Universal Dependencies\footnote{\url{http://universaldependencies.org}} (UD), and in the case of several treebanks for one language, we only use the canonical treebank. There are in total 22 treebanks, as in \newcite{Plank:2016}.\footnote{We use all training data for Czech, while \newcite{Plank:2016} only use a subset.}
Each treebank splits into train, dev, and test sets, we use the dev sets for early stop, and test on the test sets.

\subsection{Tasks}

\enote{Note}{I think you should add some motivation why three tasks and not one. Something like:}

We evaluate our method on three tagging tasks: POS tagging (\textbf{\upos{}}), morphological tagging (\textbf{\morph{}}) and supertagging (\textbf{\stag{}}).

\enote{Note}{If the reason is the tag set size then you should have those sizes somewhere. }

For POS tagging we use Universal POS tags, which is an extension of \newcite{Petrov:2012}. The universal tag set tries to capture the ``universal'' properties of words and facilitate cross-lingual learning. Therefore the tag set is very coarse and leaves out most of the language-specific properties to morphological features.

Morphological tags encode the language-specific morphological features of the words, e.g., number, gender, case. They are represented in the UD treebanks as one string  which contains several key-value pairs of morphological features.\footnote{German, French and Indonesian do not have \morph{} tags in UD-1.2, thus not evaluated in this task.}

Supertags \cite{Joshi:1994} are tags that encode more syntactic information than standard POS tags, e.g. the head direction or the subcategorization frame. We use dependency-based supertags \cite{FothBM06} which are extracted from the dependency treebanks. Adding such tags into feature models of statistical dependency parsers significantly improves their performance \cite{Ouchi:2014,Falenska:2015}. Supertags can be designed with different levels of granularity. 
We use the standard Model~1 from \newcite{Ouchi:2014}, where each tag consists of head direction, dependency label and dependent direction.
Even with the basic supertag model, the \stag{} task is more difficult than \upos{} and \morph{} because it generally requires taking long-distance dependencies between words into consideration.

We select these tasks as examples for tagging applications because they differ strongly in tag set sizes. 
Generally, the \upos{} set sizes for all the languages are no more than 17 and \stag{} set sizes are around 200. 
When treating morphological features as a string (i.e. not splitting into key-value pairs), the sizes of the \morph{} tag sets range from about 100 up to 2000.

\subsection{Setups}
As baselines to our models, we take the two state-of-the-art taggers MarMot\footnote{\url{http://cistern.cis.lmu.de/marmot/}} (denoted as CRF) and bilstm-aux\footnote{\url{https://github.com/bplank/bilstm-aux}} (denoted as LSTM). We train the taggers with the recommended hyper-parameters from the documentation.

To ensure a fair comparison (especially between LSTM and CNN), we generally treat the three tasks equally, and do not apply task-specific tuning on them, i.e., using the same features and same model hyper-parameters in each single task. Also, we do not use any pre-trained word embeddings.

For the LSTM tagger, we use the recommended hyper-parameters in the documentation\footnote{We use the most recent version of the tagger and stacking 3 layers of LSTM as recommended. The average accuracy for \upos{} in our evaluation is slightly lower than reported in the paper, presumably because of different versions of the tagger, but it does not influence the conclusion.} 
including 64-dimensional word embeddings ($\vec{w}$) and 100-dimensional composed vectors ($\vec{c}$). 
We train the $\vec{w}$, $\vec{c}$ and $\vec{w}+\vec{c}$ models as in \newcite{Plank:2016}. We train the CNN taggers with the same dimensionalities for word representations.


For the CRF tagger, we predict \upos{} and \morph{} jointly as in the standard setting for MarMot, which performs much better than with separate predictions, as shown in \newcite{Mueller:2013} and in our preliminary experiments. Also, it splits the morphological tags into key-value pairs, whereas the neural taggers treat the whole string as a tag.\footnote{Since we use the CRF tagger as a non-neural baseline model, we prefer to use the settings which maximize its performances than the rigorously equal but suboptimal settings. }
 We predict \stag{} as a separate task. 

\subsection{Results}
  The test results for the three tasks are shown in Table~\ref{tbl:results} in three groups. The first group of seven columns are the results for \upos{}, where both LSTM and CNN have three variations of input features: word only ($\vec{w}$), character only ($\vec{c}$) and both ($\vec{w}+\vec{c}$). For \morph{} and \stag{}, we only use the $\vec{w}+\vec{c}$ setting for both LSTM and CNN.

  On macro-average, three taggers perform close in the \upos{} task, with the CNN tagger being slightly better.
  In the \morph{} task, CNN is again slightly ahead of CRF, while LSTM is about 2 points behind. 
  In the \stag{} task, CNN outperforms both taggers by a large margin: 2 points higher than LSTM and 8 points higher than CRF. 

  While considering the input features of the LSTM and CNN taggers, both taggers perform close with only $\vec{w}$ as input, which suggests that the two taggers are comparable in encoding context for tagging \upos. However, with only $\vec{c}$, CNN performs much better than LSTM (95.54 vs. 92.61), and close to $\vec{w}+\vec{c}$ (96.18). Also, $\vec{c}$ consistently outperforms $\vec{w}$ for all languages. This suggests that the CNN model alone is capable of learning most of the information that the word-level model can learn, while the LSTM model is not.

  The more interesting cases are \morph{} and \stag{}, where CNN performs much higher than LSTM. 
  We hypothesize three possible reasons to explain the considerably large difference. 
  First, the LSTM tagger may be more sensitive to hyper-parameters and requires task specific tuning. We use the same setting which is tuned for the \upos{} task, thus it underperforms in the other tasks.
  Second, the LSTM tagger may not deal well with large tag sets. The tag set size for \morph{} are larger than \upos{} in orders of magnitudes, especially for Czech, Basque, Finnish and Slovene, all of which have more than 1000 distinct \morph{} tags in the training data, and the LSTM performs poorly on these languages.
  Third, the LSTM has theoretically unlimited access to all the tokens in the sentence, but in practice it might not learn the context as good as the CNN. In the LSTM model, the information of long-distance contexts will gradually fade away during the recurrence, whereas in the CNN model, all words are treated equally as long as they are in the context window.
  Therefore the LSTM underperforms in the \stag{} task, where the information from long-distance context is more important.


\begin{table*}[!h]
\footnotesize
\centering
\begin{tabular}{l | c : c c c : c c c | c c c | c c c }
\toprule
& \multicolumn{7}{|c|}{\textbf{\upos{}}} & \multicolumn{3}{c|}{\textbf{\morph{}}} & \multicolumn{3}{c}{\textbf{\stag{}}} \\
& CRF & \multicolumn{3}{c:}{LSTM} & \multicolumn{3}{c|}{CNN}  & {CRF} & {LSTM} & {CNN}  & {CRF} & {LSTM} & {CNN} \\
& & $\vec{w}$ & $\vec{c}$ & $\vec{w}+\vec{c}$ & $\vec{w}$ & $\vec{c}$ & $\vec{w}+\vec{c}$ &&$\vec{w}+\vec{c}$ &$\vec{w}+\vec{c}$&&$\vec{w}+\vec{c}$ & $\vec{w}+\vec{c}$ \\
\midrule
avg & 96.02 & 92.26 & 92.61 & 95.82 & 92.65 & 95.54 & \textbf{96.18} & 93.72 & 91.62 & \textbf{93.90} & 76.10 & 82.50 & \textbf{84.69} \\
\midrule
ar  & 98.83 & 95.05 & 98.35 & 98.88 & 95.30 & 98.89 & \textbf{99.00} & 98.11 & 97.91 & \textbf{98.45} & 79.67 & 83.70 & \textbf{85.51}\\
bg  & 98.11 & 94.96 & 96.94 & 98.07 & 95.25 & 97.79 & \textbf{98.20} & \textbf{95.12} & 92.28 & 94.85 & 78.91 & 85.91 & \textbf{87.64}\\
cs  & 98.74 & 96.12 & 92.98 & 98.40 & 96.36 & 98.35 & \textbf{98.79} & 93.81 & 90.21 & \textbf{94.45} & 76.33 & 81.43 & \textbf{87.46}\\
da  & 95.96 & 91.74 & 94.29 & \textbf{96.06} & 92.08 & 95.24 & 95.92 & \textbf{95.50} & 94.15 & 95.14 & 73.83 & 81.00 & \textbf{81.82}\\
de  & \textbf{92.77} & 89.91 & 88.97 & 92.57 & 90.21 & 92.44 & 92.73 & - & - & - & 70.56 & 77.58 & \textbf{79.69}\\
en  & 94.49 & 91.58 & 88.99 & 94.17 & 92.64 & 93.76 & \textbf{94.76} & 95.69 & 95.45 & \textbf{95.88} & 75.57 & 83.27 & \textbf{85.87}\\
es  & 95.28 & 93.27 & 91.41 & 94.62 & 93.95 & 95.36 & \textbf{95.65} & 96.14 & 95.26 & \textbf{96.34} & 78.07 & 83.80 & \textbf{86.27}\\
eu  & 94.79 & 88.70 & 89.80 & 93.99 & 89.69 & 94.31 & \textbf{94.94} & \textbf{89.60} & 84.32 & 89.06 & 70.44 & 77.88 & \textbf{80.43}\\
fa  & 96.82 & 95.67 & 94.73 & 96.95 & 95.97 & 96.12 & \textbf{97.12} & \textbf{96.56} & 96.37 & 96.50 & 76.76 & 83.21 & \textbf{83.25}\\
fi  & \textbf{95.79} & 87.78 & 84.41 & 94.16 & 88.24 & 94.33 & 95.31 & \textbf{94.33} & 87.33 & 93.82 & 70.69 & 76.65 & \textbf{82.63}\\
fr  & 95.98 & 94.34 & 91.82 & 95.85 & 94.56 & 95.68 & \textbf{96.27} & - & - & - & 78.36 & 84.01 & \textbf{85.44}\\
he  & 95.48 & 93.81 & 92.96 & 95.62 & 93.81 & 94.68 & \textbf{96.04} & 92.92 & 91.27 & \textbf{93.29} & 76.73 & 82.56 & \textbf{85.44}\\
hi  & 96.36 & 95.66 & 91.12 & 96.23 & 96.04 & 95.77 & \textbf{96.69} & 90.93 & 90.78 & \textbf{92.11} & 85.54 & 89.62 & \textbf{90.08}\\
hr  & \textbf{95.56} & 88.10 & 94.47 & 94.69 & 88.92 & 94.76 & 95.05 & 87.25 & 84.56 & \textbf{87.73} & 71.42 & 77.77 & \textbf{79.27}\\
id  & \textbf{93.51} & 90.40 & 90.76 & 92.97 & 91.15 & 92.32 & 93.44 & - & - & - & 75.37 & 80.55 & \textbf{81.63}\\
it  & \textbf{97.74} & 96.04 & 94.64 & 97.55 & 96.54 & 97.08 & 97.62 & \textbf{97.63} & 97.13 & 97.47 & 84.02 & 89.10 & \textbf{90.89}\\
nl  & 91.03 & 85.09 & 86.52 & 92.23 & 83.74 & 92.05 & \textbf{93.11} & 92.32 & 91.26 & \textbf{93.12} & 67.04 & 77.71 & \textbf{79.68}\\
no  & 97.61 & 94.39 & 93.32 & 97.49 & 94.60 & 97.05 & \textbf{97.65} & \textbf{96.03} & 94.85 & 95.74 & 79.99 & 86.45 & \textbf{89.41}\\
pl  & \textbf{96.92} & 89.53 & 95.05 & 96.30 & 90.48 & 96.41 & 96.83 & \textbf{87.74} & 82.34 & 87.13 & 76.09 & 80.00 & \textbf{83.45}\\
pt  & \textbf{97.78} & 94.20 & 94.95 & 97.53 & 94.41 & 97.22 & 97.46 & 94.99 & 94.75 & \textbf{95.76} & 78.68 & 86.02 & \textbf{87.42}\\
sl  & 96.60 & 90.43 & 96.35 & \textbf{97.42} & 91.02 & 96.89 & 97.16 & 90.41 & 86.47 & \textbf{91.94} & 76.35 & 85.67 & \textbf{86.45}\\
sv  & 96.23 & 93.04 & 94.48 & 96.20 & 93.27 & 95.38 & \textbf{96.28} & \textbf{95.65} & 94.08 & 95.30 & 73.81 & 81.04 & \textbf{83.34}\\
\bottomrule
\end{tabular}
\caption{Tagging accuracies of the three taggers in the three tasks on the test set of UD-1.2, the highest accuracy for each task on each language is marked in bold face.}\label{tbl:results}
\end{table*}

\subsection{Unnormalized Text}

  It is a common scenario to use a model trained with news data to process text from social media, which could include intentional or unintentional misspellings. Unfortunately, we do not have social media data to test the taggers.
  However, we design an experiment to simulate unnormalized text, by systematically editing the words in the dev sets with three operations: insertion, deletion and substitution. 
  For example, if we modify a word {\em abcdef} at position 2 (0-based), the modified words would be {\em abxcdef}, {\em abdef}, and {\em abxdef}, where {\em x} is a random character from the alphabet of the language.

\begin{figure}[h]
\includegraphics[width=.48\textwidth]{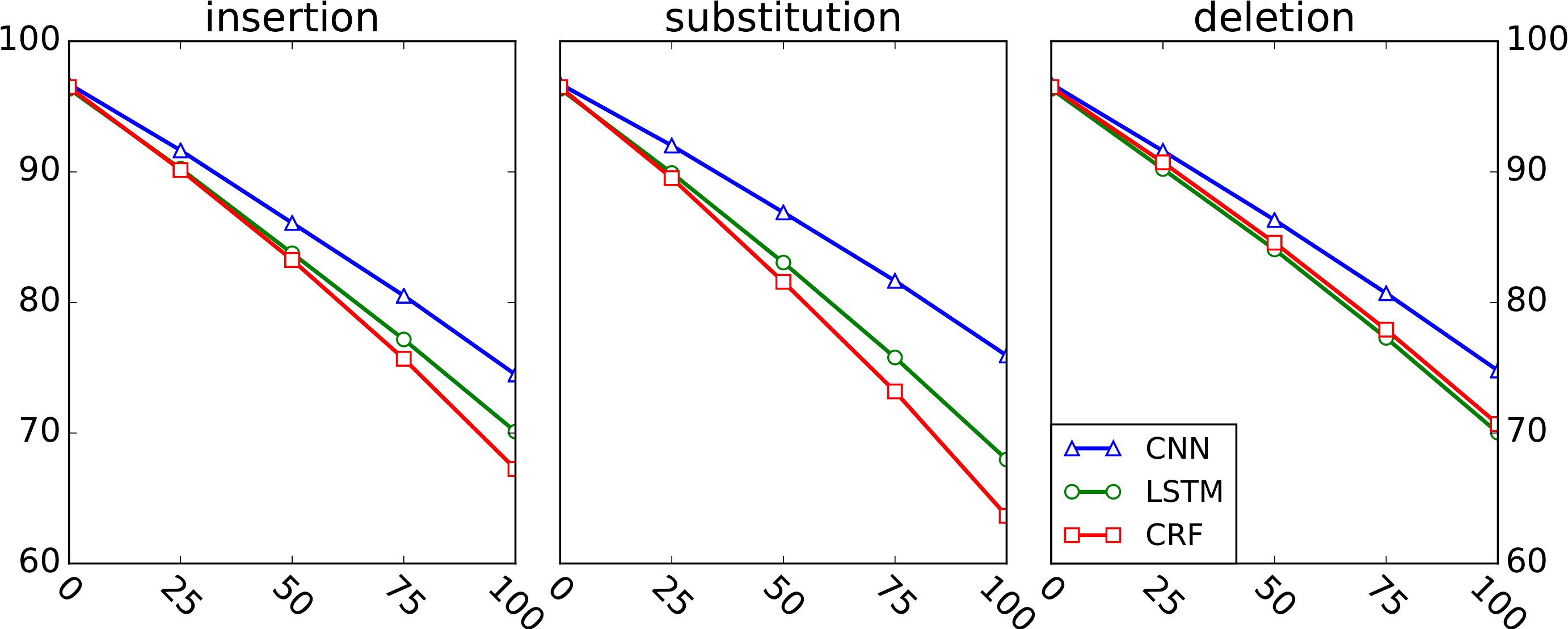}
\caption{\upos{} tagging accuracies on the dev set with the three modifications of different degrees.\label{fig:edits}}
\end{figure}

  For each operation, we create a group of modified dev sets, where all words longer than two characters are edited by the operation with a probability of 0.25, 0.5, 0.75, or 1.
  For each language, we use the models trained on the normal training sets and predict \upos{} for the three groups of modified dev set.
  The average accuracies are shown in Figure~\ref{fig:edits}.

  Generally, all models suffer from the increasing degrees of unnormalized texts, but CNN always suffers the least. In the extreme case where almost all words are unnormalized, CNN performs 4 to 8 points higher than LSTM and 4 to 11 points higher than CRF. This suggests that the CNN is more robust to misspelt words. 
  While looking into the specific cases of misspelling, CNN is more sensitive to insertion and deletion, while CRF and LSTM are more sensitive to substitution.

\section{Conclusion}
In this paper, we propose a general-purpose tagger that uses two CNNs for both character composition and context encoding. On the universal dependency treebanks (v1.2), the tagger achieves state-of-the-art results for POS tagging and morphological tagging, and to the best of our knowledge, it also performs best for supertagging. 
The tagger works well across different tagging tasks without tuning the hyper-parameters, and it is also robust against unnormalized text.





\newpage
\newpage
\bibliography{sclem2017}

\begin{thebibliography}{}
\expandafter\ifx\csname natexlab\endcsname\relax\def\natexlab#1{#1}\fi

\bibitem[{Ballesteros et~al.(2015)Ballesteros, Dyer, and
  Smith}]{Ballesteros:2015}
Miguel Ballesteros, Chris Dyer, and A.~Noah Smith. 2015.
\newblock \href{https://doi.org/10.18653/v1/D15-1041}{Improved transition-based
  parsing by modeling characters instead of words with lstms}.
\newblock In {\em Proceedings of the 2015 Conference on Empirical Methods in
  Natural Language Processing\/}. Association for Computational Linguistics,
  pages 349--359.
\newblock
  \href{https://doi.org/10.18653/v1/D15-1041}{https://doi.org/10.18653/v1/D15-1041}.

\bibitem[{Costa-juss{\`a} and Fonollosa(2016)}]{Costa:2016}
R.~Marta Costa-juss{\`a} and R.~Jos{\'e}~A. Fonollosa. 2016.
\newblock \href{https://doi.org/10.18653/v1/P16-2058}{Character-based neural
  machine translation}.
\newblock In {\em Proceedings of the 54th Annual Meeting of the Association for
  Computational Linguistics (Volume 2: Short Papers)\/}. Association for
  Computational Linguistics, pages 357--361.
\newblock
  \href{https://doi.org/10.18653/v1/P16-2058}{https://doi.org/10.18653/v1/P16-2058}.

\bibitem[{dos Santos and Guimar{\~a}es(2015)}]{Santos:2015}
Cicero dos Santos and Victor Guimar{\~a}es. 2015.
\newblock \href{https://doi.org/10.18653/v1/W15-3904}{Boosting named entity
  recognition with neural character embeddings}.
\newblock In {\em Proceedings of the Fifth Named Entity Workshop\/}.
  Association for Computational Linguistics, pages 25--33.
\newblock
  \href{https://doi.org/10.18653/v1/W15-3904}{https://doi.org/10.18653/v1/W15-3904}.

\bibitem[{dos Santos and Zadrozny(2014)}]{Santos:2014}
Cicero dos Santos and Bianca Zadrozny. 2014.
\newblock Learning character-level representations for part-of-speech tagging.
\newblock In {\em Proceedings of the 31st International Conference on Machine
  Learning (ICML-14)\/}. pages 1818--1826.

\bibitem[{Fale\'{n}ska et~al.(2015)Fale\'{n}ska, Bj\"{o}rkelund,
  \c{C}etino\u{g}lu, and Seeker}]{Falenska:2015}
Agnieszka Fale\'{n}ska, Anders Bj\"{o}rkelund, \"{O}zlem \c{C}etino\u{g}lu, and
  Wolfgang Seeker. 2015.
\newblock \href{http://www.aclweb.org/anthology/W15-2215}{Stacking or
  supertagging for dependency parsing -- what's the difference?}
\newblock In {\em Proceedings of the 14th International Conference on Parsing
  Technologies\/}. Association for Computational Linguistics, Bilbao, Spain,
  pages 118--129.
\newblock
  \href{http://www.aclweb.org/anthology/W15-2215}{http://www.aclweb.org/anthology/W15-2215}.

\bibitem[{Foth et~al.(2006)Foth, By, and Menzel}]{FothBM06}
Kilian~A. Foth, Tomas By, and Wolfgang Menzel. 2006.
\newblock \href{http://aclweb.org/anthology/P06-1037}{Guiding a constraint
  dependency parser with supertags}.
\newblock In {\em {ACL} 2006, 21st International Conference on Computational
  Linguistics and 44th Annual Meeting of the Association for Computational
  Linguistics, Proceedings of the Conference, Sydney, Australia, 17-21 July
  2006\/}.
\newblock
  \href{http://aclweb.org/anthology/P06-1037}{http://aclweb.org/anthology/P06-1037}.

\bibitem[{Gehring et~al.(2017)Gehring, Auli, Grangier, Yarats, and
  Dauphin}]{Gehring:2017}
Jonas Gehring, Michael Auli, David Grangier, Denis Yarats, and Yann~N Dauphin.
  2017.
\newblock Convolutional sequence to sequence learning.
\newblock {\em arXiv preprint arXiv:1705.03122\/} .

\bibitem[{Joshi and Bangalore(1994)}]{Joshi:1994}
Aravind~K. Joshi and Srinivas Bangalore. 1994.
\newblock \href{https://doi.org/10.3115/991886.991912}{{Disambiguation of Super
  Parts of Speech (or Supertags): Almost Parsing}}.
\newblock In {\em Proceedings of the 15th Conference on Computational
  Linguistics - Volume 1\/}. Association for Computational Linguistics,
  Stroudsburg, PA, USA, COLING '94, pages 154--160.
\newblock
  \href{https://doi.org/10.3115/991886.991912}{https://doi.org/10.3115/991886.991912}.

\bibitem[{Kim et~al.(2016)Kim, Jernite, Sontag, and Rush}]{Kim:2016}
Yoon Kim, Yacine Jernite, David Sontag, and Alexander~M. Rush. 2016.
\newblock Character-aware neural language models.
\newblock In {\em Proceedings of the Thirtieth {AAAI} Conference on Artificial
  Intelligence, February 12-17, 2016, Phoenix, Arizona, {USA.}\/}. {AAAI}
  Press, pages 2741--2749.

\bibitem[{Ling et~al.(2015)Ling, Dyer, Black, Trancoso, Fermandez, Amir,
  Marujo, and Luis}]{Ling:2015}
Wang Ling, Chris Dyer, W.~Alan Black, Isabel Trancoso, Ramon Fermandez, Silvio
  Amir, Luis Marujo, and Tiago Luis. 2015.
\newblock \href{https://doi.org/10.18653/v1/D15-1176}{Finding function in form:
  Compositional character models for open vocabulary word representation}.
\newblock In {\em Proceedings of the 2015 Conference on Empirical Methods in
  Natural Language Processing\/}. Association for Computational Linguistics,
  pages 1520--1530.
\newblock
  \href{https://doi.org/10.18653/v1/D15-1176}{https://doi.org/10.18653/v1/D15-1176}.

\bibitem[{M\"uller et~al.(2013)M\"uller, Schmid, and Sch\"utze}]{Mueller:2013}
Thomas M\"uller, Helmut Schmid, and Hinrich Sch\"utze. 2013.
\newblock {Efficient Higher-Order {CRF}s for Morphological Tagging}.
\newblock In {\em In Proceedings of EMNLP\/}.

\bibitem[{Ouchi et~al.(2014)Ouchi, Duh, and Matsumoto}]{Ouchi:2014}
Hiroki Ouchi, Kevin Duh, and Yuji Matsumoto. 2014.
\newblock \href{http://www.aclweb.org/anthology/E14-4030}{{Improving Dependency
  Parsers with Supertags}}.
\newblock In {\em Proceedings of the 14th Conference of the European Chapter of
  the Association for Computational Linguistics, volume 2: Short Papers\/}.
  Association for Computational Linguistics, Gothenburg, Sweden, pages
  154--158.
\newblock
  \href{http://www.aclweb.org/anthology/E14-4030}{http://www.aclweb.org/anthology/E14-4030}.

\bibitem[{Petrov et~al.(2012)Petrov, Das, and McDonald}]{Petrov:2012}
Slav Petrov, Dipanjan Das, and Ryan McDonald. 2012.
\newblock A universal part-of-speech tagset.
\newblock In {\em Proceedings of the Eight International Conference on Language
  Resources and Evaluation (LREC'12)\/}. European Language Resources
  Association (ELRA), Istanbul, Turkey.

\bibitem[{Plank et~al.(2016)Plank, S{\o}gaard, and Goldberg}]{Plank:2016}
Barbara Plank, Anders S{\o}gaard, and Yoav Goldberg. 2016.
\newblock \href{https://doi.org/10.18653/v1/P16-2067}{Multilingual
  part-of-speech tagging with bidirectional long short-term memory models and
  auxiliary loss}.
\newblock In {\em Proceedings of the 54th Annual Meeting of the Association for
  Computational Linguistics (Volume 2: Short Papers)\/}. Association for
  Computational Linguistics, pages 412--418.
\newblock
  \href{https://doi.org/10.18653/v1/P16-2067}{https://doi.org/10.18653/v1/P16-2067}.

\bibitem[{Vu et~al.(2016)Vu, Adel, Gupta, and Sch{\"u}tze}]{Vu:2016}
Thang~Ngoc Vu, Heike Adel, Pankaj Gupta, and Hinrich Sch{\"u}tze. 2016.
\newblock \href{https://doi.org/10.18653/v1/N16-1065}{Combining recurrent and
  convolutional neural networks for relation classification}.
\newblock In {\em Proceedings of the 2016 Conference of the North American
  Chapter of the Association for Computational Linguistics: Human Language
  Technologies\/}. Association for Computational Linguistics, pages 534--539.
\newblock
  \href{https://doi.org/10.18653/v1/N16-1065}{https://doi.org/10.18653/v1/N16-1065}.

\bibitem[{Yu and Vu(2017)}]{Yu:2017}
Xiang Yu and Ngoc~Thang Vu. 2017.
\newblock Character composition model with convolutional neural networks for
  dependency parsing on morphologically rich languages.
\newblock {\em arXiv preprint arXiv:1705.10814\/} .

\end{thebibliography}
\bibliographystyle{emnlp_natbib}

\end{document}